\pdfoutput=1
\documentclass[11pt]{article}

\usepackage[final]{acl}

\usepackage{times}
\usepackage{latexsym}

\usepackage[T1]{fontenc}
\usepackage[utf8]{inputenc}

\usepackage{microtype}

\usepackage{inconsolata}

\usepackage{graphicx}
\usepackage{booktabs}
\usepackage{comment}
\usepackage{listings}
\title{Scaling Context, Not Parameters: Training a Compact 7B Language Model for Efficient Long-Context Processing}

\author{Chen Wu \\
  Amazon Web Services \\
  \texttt{wuc@amazon.com} \\\And
  Yin Song \\
  Amazon Web Services \\
  \texttt{yinsong@amazon.com} \\}

\begin{document}
\maketitle
\begin{abstract}
We present MegaBeam-Mistral-7B\footnote{\url{https://huggingface.co/aws-prototyping/MegaBeam-Mistral-7B-512k}}, a language model that supports 512K-token context length. Our work addresses practical limitations in long-context training, supporting real-world tasks such as compliance monitoring and verification. Evaluated on three long-context benchmarks, our 7B-parameter model demonstrates superior in-context learning performance on HELMET and robust retrieval and tracing capability on RULER. It is currently the only open model to achieve competitive long-range reasoning on BABILong at 512K context length without RAG or targeted fine-tuning. Released as fully open source under the Apache 2.0 license, the model has been downloaded over 100,000 times on Hugging Face. 
\end{abstract}

\section{Introduction}

MegaBeam-Mistral-7B is a compact 7B-parameter language model capable of processing sequences with half-a-million tokens. Developed with customer engagements in mind, we thoroughly evaluated its long-context capabilities across multiple benchmarks. 

MegaBeam delivers strong performance across three key long-context benchmarks. On RULER at 128K context length, it outperforms both GPT-4-1106 and larger open-source models like Llama-3.1-70B. On BABILong at 64K context, it achieves 48.2\% accuracy—comparable to models with 8x more parameters. On HELMET, it attains a leading 85\% in-context learning score at 128K tokens. Significantly, MegaBeam achieves a competitive 35\% score on 512K-token BABILong tasks without RAG or task-specific tuning, making it the only open model to effectively utilise such extreme context lengths for solving novel reasoning tasks.

MegaBeam's development was shaped primarily by our engagements with customers across diverse sectors, including digital design, banking, life sciences, and GenAI native startups. 

For example, large enterprises face daily challenges in verifying compliance across their customer interactions, which often involve processing lengthy conversation transcripts and transaction logs. 
To tackle this challenge, we deployed MegaBeam as a prototype compliance verification solution, performing three key functions: First, it identifies and matches specific sections of customer interactions with relevant Standard Operating Procedures guidelines. It then classifies these matched segments for compliance adherence, examining elements such as required disclosures, proper documentation, and procedural steps. Finally, it provides detailed reasoning for each compliance assessment by comparing the actual interaction patterns against mandated procedures.
The ability to digest customer interaction logs alongside SOPs within its context eliminates the need to chunk conversations. MegaBeam enables efficient compliance monitoring by maintaining the complete context of customer interactions alongside regulatory requirements.

The following sections detail our technical approach to achieving these capabilities, addressing challenges in training methodology and system-level optimisations required for robust performance in production environments.

\section{Related Work}

%\textbf{Long Context Models and Training.} 
Recent advances in LLM context length extension have emerged through improved training methodologies. MiniCPM \citep{hu2024minicpm} and Yi \citep{young2024yi} demonstrated that even smaller models could handle 200K+ contexts through targeted training approaches. \citet{fu2024data} established that modest amounts of long-sequence text (1-2B tokens) can effectively extend context capabilities without full retraining. To address computational challenges, sequence parallel techniques such as Ring Attention \citep{liu2023ring} and DeepSpeed-Ulysses \citep{jacobs2023deepspeed} have made training with extremely long sequences more feasible.

%\textbf{Evaluation Frameworks.}
Several long-context benchmarks have emerged to systematically evaluate long-context capabilities. RULER \citep{hsieh2024ruler} focuses on retrieval and multi-hop reasoning, BABILong \citep{kuratov2024babilong} tests reasoning over extremely long documents, and HELMET \citep{yen2024helmet} provides application-centric metrics across diverse downstream tasks.

%\textbf{Position Encoding.} 
Adjusting the theta base parameter in Rotary Position Encoding (RoPE) \citep{su2024roformer} has emerged as the dominant approach for extending context length. Recent theoretical work by \citet{xu2024base} has established lower bounds for effective theta values based on target sequence lengths. LongRoPE \citep{ding2024longrope} introduced innovative position encoding modifications, enabling models to handle substantially longer sequences with minimal additional training.

Our work builds upon these foundations, focusing specifically on efficient training techniques that allow smaller models (7B parameters) to handle extremely long contexts (512K tokens), previously thought to require substantially larger models or computational resources.

\section{Training}
\label{sect:training}

The training methodology for MegaBeam builds upon key insights from several previous studies. Drawing from \cite{young2024yi} and \cite{fu2024data}, we implemented lightweight continual pretraining with long-context data, confirming that $\leq2$B tokens are sufficient for extending context length capabilities. We also incorporated findings from the MiniCPM model \citep{hu2024minicpm} regarding the optimal balance between short and long training examples—specifically their discovery that mixing ratios are crucial for maintaining performance across different context lengths.

\begin{figure*}[!t]
\begin{center}
\includegraphics[scale=0.2]{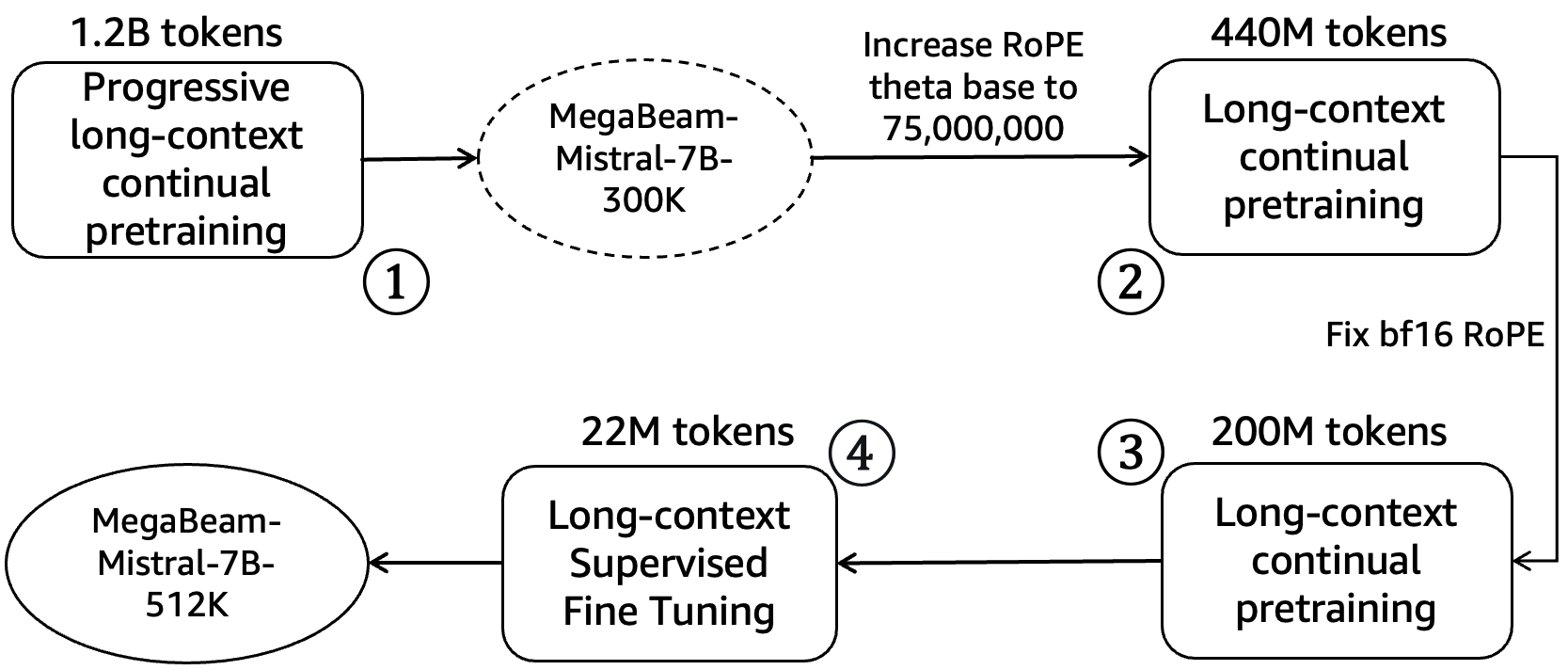}
\end{center}
\caption{Overview of MegaBeam's training methodology: four sequential phases}
\label{fig:training-approach}
\end{figure*}

The training process consists of four phases (Fig \ref{fig:training-approach}) with varying token counts and sequence lengths. Using Mistral-7B-Instruct-v0.2 \citep{mistral2023_7b} as the base model, the first phase involved progressive long-context training on 1.2B tokens of organically long documents from diverse sources: source code (70\%), research papers (10\%), open web content (15\%), and public domain books (5\%). This initial phase processed 0.64B tokens as 300K-token sequences and 0.56B tokens as 600K-token sequences. Although we trained with sequence lengths up to 600K tokens, our evaluation using the Needle-in-a-Haystack (NIAH) benchmark \citep{niah2024_code} revealed significant performance degradation when processing sequences longer than 300K tokens. We named this intermediate checkpoint \texttt{MegaBeam-Mistral-7B-300K} to reflect its effective context length. 

To address the performance degradation beyond 300K tokens, we increased the RoPE theta base from $25\_000\_000$ to $75\_000\_000$ and trained on an additional 0.18B tokens using 600K-token sequences. This improved overall long-context performance but led to poor NIAH scores at sequence endpoints (depth 0 and 100). We attributed this to insufficient training on shorter sequences with the new RoPE configuration -- a hypothesis confirmed when additional training on 0.26B tokens of shorter sequences (32K-80K) resolved the endpoint issues while maintaining long-sequence performance.

After addressing a critical numerical precision issue in the bfloat16 RoPE implementation, we conducted a third round of long-context continual pretraining using 0.2B tokens. The training data was distributed across different sequence lengths: 1,200 sequences of 80K tokens (96M total), 300 sequences of 256K tokens (77M total), and 30 sequences of 512K tokens (15M total). This balanced distribution ensured robust performance across all context windows.

The final phase involved long-context supervised fine-tuning (SFT) on a small 22M-token data set, producing \texttt{MegaBeam-Mistral-7B-512K}. Following insights from \cite{hu2024minicpm} and \cite{young2024yi}, we created synthetic documents (64K-512K tokens) by restructuring real question-answer pairs to specifically challenge long-range information retrieval.

This phased approach combines planned length progression with solutions to unexpected challenges discovered during development, enabling effective scaling to longer contexts while maintaining performance stability.

\section{Solving Practical Issues}

\subsection{RoPE theta base}
As discussed in Section \ref{sect:training}, we tuned the RoPE theta base through progressive pretraining to improve NIAH benchmark performance. Our experimentally determined values—$25\_000\_000$ for sequences of 256K tokens and $75\_000\_000$ for sequences of 512K tokens—closely match the theoretical lower bounds derived by \cite{xu2024base}: $\beta=0.0424L^{1.628}$, which yields $28\_000\_000$ and $86\_000\_000$ respectively. 

Our experiments also revealed additional insights. Specifically, setting the base to $100\_000\_000$ systematically degraded performance at the sequence endpoints (depth 0 and 100) for long sequences. This observation seems to align with \citep{liu2023scaling}. When the base value substantially exceeds the lower bound, it creates positional embeddings with wavelengths longer than the training context length. This means some dimensions cannot complete a full $2\pi$ rotation during training, potentially leading to hallucinations during inference.

\begin{figure*}[!t]
\begin{center}
\includegraphics[scale=0.5]{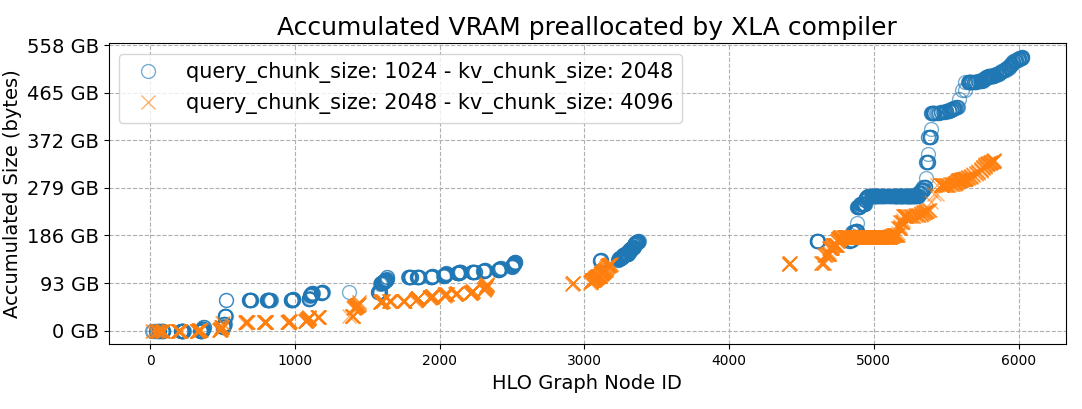}
\end{center}
\caption{Accumulated memory pre-allocation by XLA compiler under two chunk size configurations. The orange line (larger chunks) demonstrates reduced memory footprint compared to the blue line (smaller chunks) throughout the HLO graph, with peak memory reduction of 186GB.}
\label{fig:mem_pre_aloc}
\end{figure*}

\subsection{bf16 and RoPE}
We encountered recall failures in NIAH benchmark. Specifically, when processing longer contexts, the model consistently dropped the last one digit when recalling numbers (e.g., recalling $7418118$ as $741811$).
The root cause was traced to numerical precision limitations of bfloat16 when handling large position indices in RoPE calculations. While float32 maintains sufficient precision across all position indices, bfloat16's reduced mantissa bits lead to significant precision loss when representing large positions, despite having comparable range to float32. This precision loss directly impacts RoPE's ability to accurately encode positional information for tokens far into a long sequence.

The solution involves disabling \texttt{autocast} and forcing float32 precision specifically for the critical RoPE calculations while maintaining bfloat16 for the rest of the model operations. This targeted precision management ensures accurate positional encoding while retaining the memory and computational benefits of bfloat16 for other operations. This fix was crucial for enabling reliable long-context processing in MegaBeam. After we have released MegaBeam, a comprehensive analysis of this precision-related issue was later discussed in \cite{wang2024precision}.

\subsection{Ring Attention}
Ring Attention \citep{liu2023ring} is an effective Sequence Parallel (SP) mechanism for distributed long sequence training. It organises accelerators in a ring topology where attention keys and values rotate in a peer-to-peer fashion between devices while queries remain fixed on their assigned devices. 

There are alternative approaches to SP besides Ring Attention, such as DeepSpeed-Ulysses \citep{jacobs2023deepspeed}. However, DeepSpeed-Ulysses requires all-to-all collective communication to transpose partitions from sequence to head dimensions, and each device must store a complete KV head for the entire sequence length. As a result, its degree of sequence parallelism (DoSP) is constrained by the number of KV heads. Ring Attention, in contrast, allows DoSP to scale linearly with the total number of available devices. These advantages led us to adopt the JAX-based \citep{lwm2024_code} Ring Attention implementation for our sequence parallelism. 

Although the JAX codebase \citep{lwm2024_code} supports interleaving Tensor Parallelism (TP) with SP, we disable TP (setting it to 1) for sequences longer than 64K tokens. This prioritisation of SP over TP allocates more VRAM to sequence parallelism, which becomes crucial as sequence lengths are growing. For larger models like 70B parameters, the optimal parallel mesh configuration between SP and TP would need to be re-established through similar experimentation. This parallelism strategy is necessary because, as demonstrated in the Megatron context parallelism example \citep{megatraon_doc}, SP and TP share a fixed pool of GPUs. Additionally, interleaving TP and SP incurs communication overhead through extra operations such as \texttt{All-Gathers} and \texttt{Reduce-Scatters}.

\subsection{XLA compiler}
\citet{liu2023ring} documented resource demands of long-context training. For sequences of 512K tokens, they had to use $16\times$A100 (80GB VRAM) to train a 7B model. We verified this limitation using their JAX codebase \citep{lwm2024_code} --- attempting to train 512K-token sequences on $8\times$A100 GPUs resulted in compilation-time OOM exceptions.

To overcome this limitation, we examined the compilation process in detail. The XLA compiler transforms JAX operations to High-Level Operations (HLO) IR, from which we identified some operation that pre-allocates 32 GB memory during compilation. Namely, the \texttt{dynamic\_update\_slice} HLO operation (shown in Appendix \ref{sec:appendix}) uses \texttt{int32} type for both input and output tensors, with the output tensor size reaching 32 GB. For our 524,288-token sequences, 8-way partitioning assigns $65,536$ tokens per GPU device. Each device's partition is then processed using 64 query chunks ($65,536/1,024$ tokens per chunk) and 32 key-value chunks ($65,536/2,048$ tokens per chunk). Based on these dimensions and the \texttt{int32} type, we hypothesise that this structure serves as a lookup table mapping QKV chunks to \texttt{segment\_ids} for intra-document attention mask generation \citep{zhao2024intradoc}.

To address this challenge, we increased both Q and K/V chunk sizes. This solution appears counter-intuitive since larger attention chunks traditionally consume more GPU HBM, as evidenced in both Block-wise Attention \cite{liu2023blockwise} (with larger blocks) and Flash Attention \cite{dao2022flashattention} (with larger tiles). However, increasing chunk sizes actually reduces the number of chunks needed, thereby decreasing the dimension extent of the lookup table tensor. This leads to reduced memory usage, contrary to conventional wisdom about chunk size and memory footprint. 

We experimented with increasing query chunks from 1024 to 2048 tokens, and key/value chunks from 2048 to 4096 tokens. Fig \ref{fig:mem_pre_aloc} compares the memory pre-allocated by the XLA compiler under these two configurations. The larger chunk sizes (orange line) consistently require less pre-allocated memory than smaller chunks (blue line) across all HLO graph nodes. This difference becomes especially significant in the later stages of the HLO graph (nodes 4000-6000). 

Most importantly, this method doubles the training context length on a single \texttt{p4de.24x} node (8x A100 with 80GB VRAM) from 256K to 512K tokens. However, while effective, this solution serves as an interim workaround. Specifically, the root issue stems from the XLA compiler materialising the massive chunk-to-segment mapping table statically. A proper solution would improve the compiler to generate dynamic mapping code, aligning with the chunked attention design.

\section{Evaluation}
\label{sect:eval}

\begin{figure*}[!t]
\centering
\includegraphics[scale=0.37]{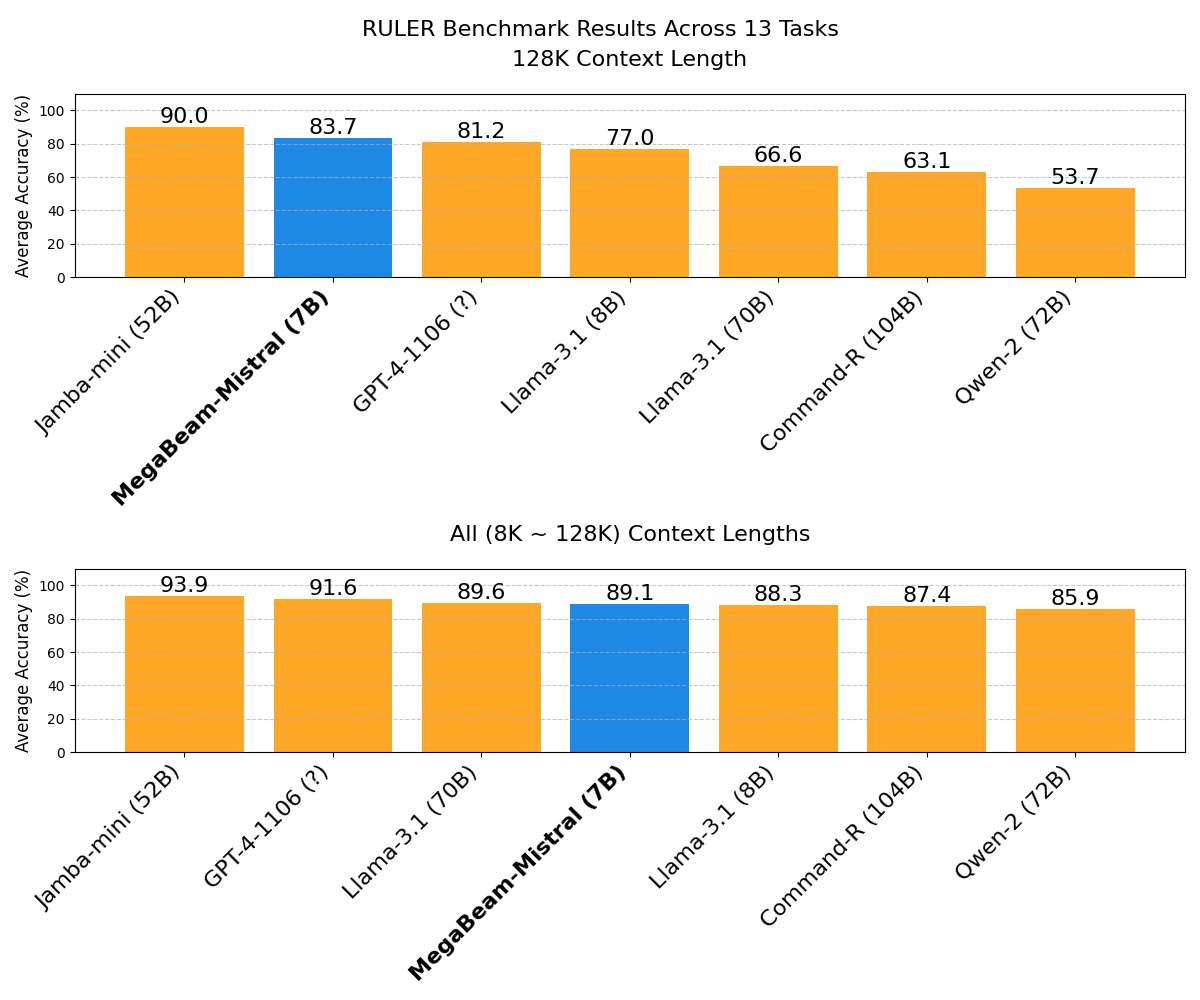}
\caption{Model performance comparison on RULER benchmark: top shows 128K context length results, bottom shows average performance across context lengths from 8K to 128K.}
\label{fig:mb-ruler}
\end{figure*}

The RULER benchmark \citep{hsieh2024ruler} specifically assesses long context capabilities in retrieval, multi-hop tracing, aggregation, and long-form question and answering. Fig.~\ref{fig:mb-ruler} shows that MegaBeam performs better than \textsc{GPT-4-1106} on the RULER benchmark when the context length is 128K. For the average performance across all lengths (8K to 128K), MegaBeam as a 7B model performs nearly on par with Llama-3.1-70B, and is ranked higher than larger models such as Llama-3.1-8B, Command-R-104B, and Qwen-2-72B. For example, MegaBeam achieves near-perfect performance on retrieval tasks (97\% on 7 out of 8 tasks at 128K), strong results on multi-hop tracing (89\% at 128K), and competitive QA performance (77.4\% on QA\_1 at 128K).

The RULER benchmark \citep{hsieh2024ruler} demonstrates that MegaBeam maintains the base model's strong performance on short contexts of 4K-16K tokens (92-94\% accuracy) while significantly outperforming Mistral-7B-Instruct-v0.2 on longer contexts (84\% vs 14\% at 128K tokens). This confirms our training approach effectively extends context length without compromising short-context capabilities.

\begin{figure*}[!t]
  \centering
  \includegraphics[scale=0.4]{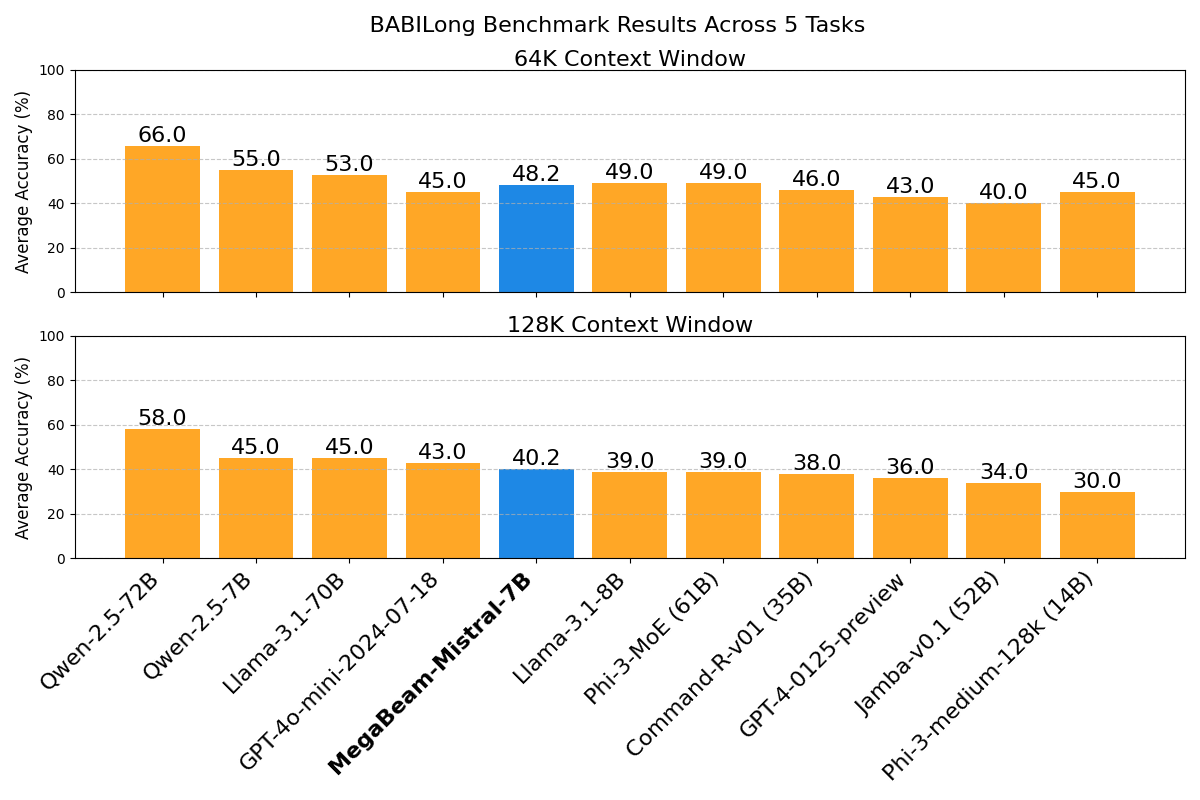}
  \caption{Performance comparison on BABILong benchmark at 64K and 128K context lengths}
  \label{fig:mb-babilong-01}
\end{figure*}

\begin{figure*}[!t]
\centering
\includegraphics[scale=0.35]{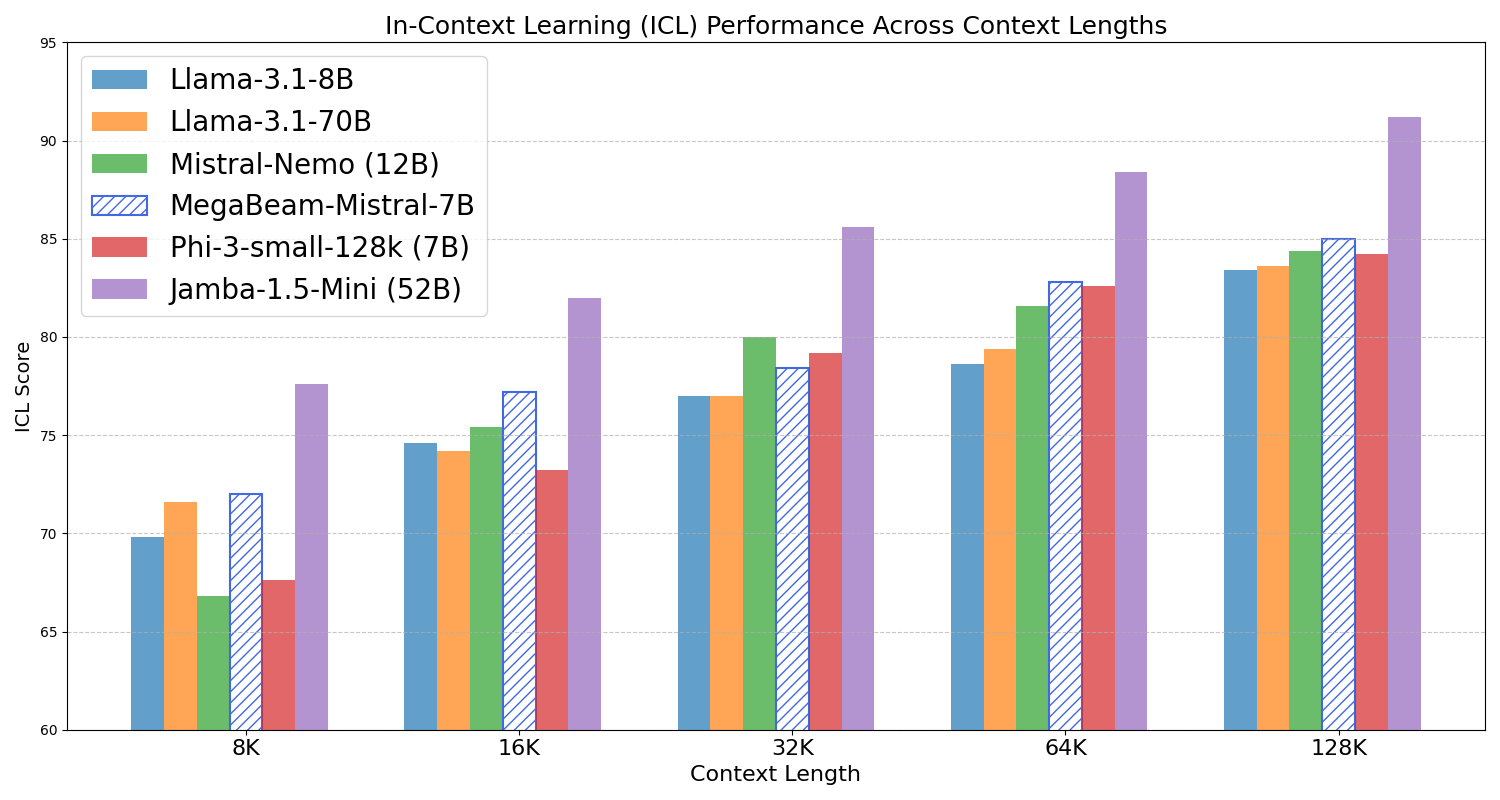}
\caption{In-Context Learning performance comparison on HELMET, showing MegaBeam's leading performance across multiple context lengths}
\label{fig:mb-helmet}
\end{figure*}

Additionally, as shown in Figure ~\ref{fig:mb-ruler}, Llama-3.1-8B outperforms its 70B counterpart, suggesting that model size alone does not guarantee superior long-context processing. In contrast, the relationship differs on BABILong, where Qwen-2.5-72B exceeds its 7B version by 13 percentage points. These varied outcomes across benchmarks support the motivation of this paper - specialised pre-training and post-training for longer contexts can enable compact models to achieve competitive performance on many long-context tasks.

\begin{figure*}[!ht]
\begin{center}
\includegraphics[scale=0.45]{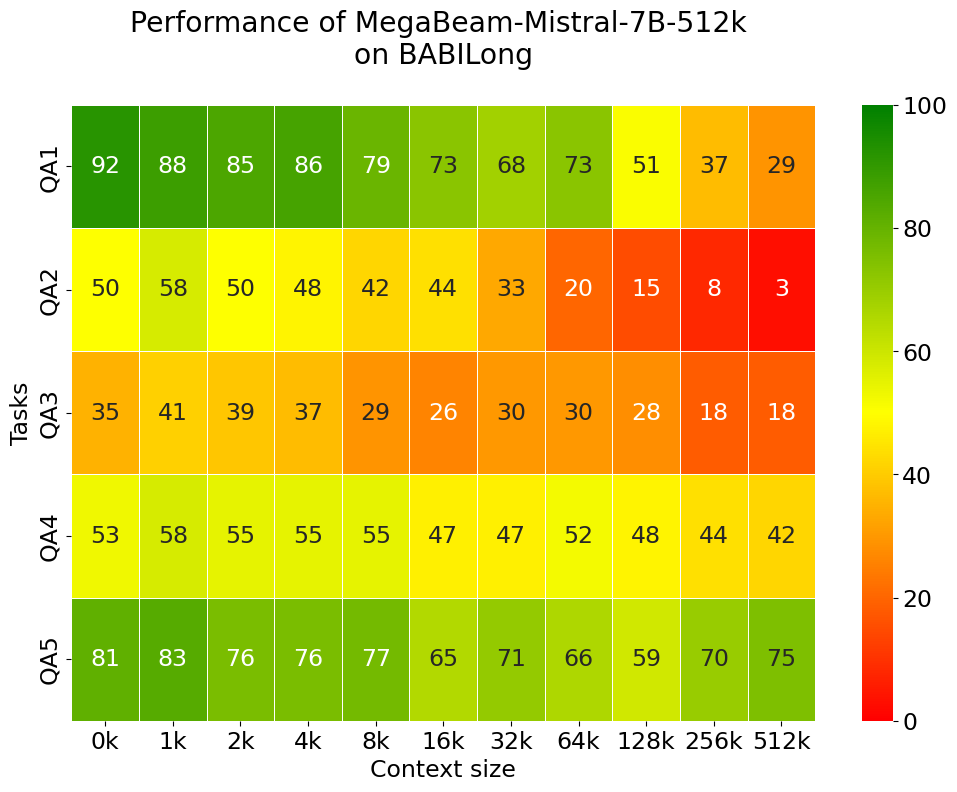}
\end{center}
\caption{Performance heatmap of MegaBeam on BABILong tasks across different context lengths (0K to 512K tokens). The model shows strong context extension capabilities on single-fact (QA1) and relational reasoning tasks (QA4, QA5), while challenges in multi-fact reasoning (QA2, QA3)}
\label{fig:babilong_all_len}
\end{figure*}

The BABILong benchmark \citep{kuratov2024babilong} evaluates the ability of LLM to perform reasoning tasks across facts distributed in extremely long documents. We conducted MegaBeam's evaluation using the official BABILong benchmark codebase\footnote{\url{https://github.com/booydar/babilong}}. Fig \ref{fig:mb-babilong-01} shows that MegaBeam achieves 48.2\% accuracy at 64K context length and 40.2\% at 128K context length, outperforming several larger models including GPT-4-0125-preview (43\% at 64K, 36\% at 128K) and matching the performance of Llama-3.1-8B and Phi-3-MoE-61B (49\% at 64K, 39\% at 128K) despite having only 7B parameters. 
MegaBeam demonstrates particularly strong performance on tasks requiring single-fact retrieval and relational reasoning, maintaining consistent performance as context length increases.
Notably, MegaBeam is currently the only open model that has achieved a competitive score (35\% as shown in Fig \ref{fig:babilong_all_len}) on the 512K context BABILong tasks without RAG or task-specific fine-tuning.

The HELMET benchmark \citep{yen2024helmet} represents the latest evaluation framework for long-context capabilities through realistic downstream tasks. It contains seven diverse, application-centric categories with model-based evaluation metrics, and few-shot prompting capabilities. Fig.~\ref{fig:mb-helmet} shows model performance comparison in the many-shot In-Context Learning (ICL) category, using performance data reported in \citep{yen2024helmet} --- At 128K context length, MegaBeam achieves an ICL score of 85\%, outperforming larger models such as Mistral-Nemo (12B), Llama-3.1 8B and 70B. 

%\begin{comment}
\section{Reasoning on BABILong}
\label{babilong}

We evaluate MegaBeam's performance on the BABILong benchmark~\citep{kuratov2024babilong}, which evaluates reasoning tasks across facts distributed in extremely long documents. As MegaBeam is fine-tuned on Mistral-7B-Instruct-v0.2 which natively supports 32K context, our analysis focuses particularly on the model's capability to extend beyond this length while maintaining performance.

MegaBeam demonstrates varying degrees of context extension capability across different tasks. For Single Supporting Fact tasks (QA1), the model maintains robust performance at 64K with 73\% accuracy, and continues to function at longer contexts with 51\% at 128K, 37\% at 256K, and 29\% at 512K. While this represents 57\% drop from 32K, the degradation is gradual and sub-linear. In Two Argument Relations tasks (QA4), MegaBeam exhibits strong stability, with performance actually improving from 47\% at 32K to 52\% at 64K, and maintaining consistent performance even at 512K (44\%), showing a high ``retention ratio" of 89\% from 32K to 512K. Similarly promising results are seen in Three Argument Relations tasks (QA5), where the model shows strong performance retention from 32K to 64K (71\% to 66\%), and maintains an even higher score at 512K (75\%), achieving an impressive 92\% retention ratio from 0K to 512K.

However, MegaBeam still faces significant challenges with multi-fact reasoning at extended contexts. In Two Supporting Facts tasks (QA2), we observe a steep performance decline from 33\% at 32K to just 3\% at 512K - a retention ratio of only 9\%. The sharp linear degradation rate suggests that our context extension approach struggles particularly with maintaining multi-fact reasoning capabilities. Similarly, Three Supporting Facts tasks (QA3) show both base model limitations (35-41\% at shorter contexts) and context extension challenges, with performance dropping to 18\% at 512K (51\% retention ratio). 

The weaker QA2/3 performance stems from multiple challenges: tracking object locations/possessions, understanding temporal order, integrating distributed information, and comprehending action-state causal relationships.

\section{Conclusion} 
We presented MegaBeam-Mistral-7B and demonstrated its competitive long-context capabilities as a smaller model trained using limited computational resources. Our work addresses key technical challenges through progressive training methods, RoPE theta tuning, position precision, and memory optimization. MegaBeam shows consistently strong performance on real-world tasks like retrieval, relation processing, and in-context learning across long contexts up to 512K tokens, while maintaining a compact model size. Its limitation in multi-hop reasoning tasks suggests areas for future improvement in both base model capabilities and context extension.

\section*{Acknowledgments}
We would like to thank three anonymous reviewers for their useful feedback to improve this paper.
% Bibliography entries for the entire Anthology, followed by custom entries
%\bibliography{anthology,custom}
% Custom bibliography entries only
\bibliography{custom}
\appendix

\section{MHLO dynamic update slice operation}
\label{sec:appendix}

\begin{figure}[!ht]
  \centering
  \begin{lstlisting}
mhlo.dynamic_update_slice(
  tensor<8x1x64x32x524288xi32>, 
  tensor<1x1x64x32x524288xi32>,  
  tensor<i32>,
  tensor<i32>,
  tensor<i32>,
  tensor<i32>,
  tensor<i32>)
  \end{lstlisting}
  %\caption{MHLO dynamic update slice operation}
  \label{lst:mhlo-dynamic-update}
\end{figure}

\end{document}